\documentclass[10pt,twocolumn]{IEEEtran}
\usepackage{graphicx}
\usepackage{amsmath}
\usepackage{latexsym}
\usepackage[table]{xcolor}
\usepackage{booktabs}
\usepackage{caption}
\usepackage{subcaption}
\usepackage{multirow}

\ifCLASSINFOpdf
\else
\fi
\begin{document}
\title{When Traffic Flow Prediction Meets Wireless Big Data Analytics}
\author{
\IEEEauthorblockN{Yuanfang Chen\IEEEauthorrefmark{1}\IEEEauthorrefmark{2}, Mohsen Guizani\IEEEauthorrefmark{3}, Yan Zhang\IEEEauthorrefmark{4}, Lei Wang\IEEEauthorrefmark{5}, Noel Crespi\IEEEauthorrefmark{1}, Gyu Myoung Lee\IEEEauthorrefmark{6}}\\
\IEEEauthorblockA{
\IEEEauthorrefmark{1}Institut Mines-T\'el\'ecom, T\'el\'ecom SudParis, France\\
\IEEEauthorrefmark{2}Department of Computer, Universit\'e Pierre et Marie CURIE, France\\
\IEEEauthorrefmark{3}Electrical and Computer Engineering, University of Idaho, USA\\
\IEEEauthorrefmark{4}University of Oslo, Norway\\
\IEEEauthorrefmark{5}School of Software, Dalian University of Technology, Dalian, China\\
\IEEEauthorrefmark{6}Liverpool John Moores University, Liverpool, UK}
}
\maketitle


\begin{abstract}
Traffic flow prediction is an important research issue for solving the traffic congestion problem in an Intelligent Transportation System (ITS).  Traffic congestion is one of the most serious problems in a city, which can be predicted in advance by analyzing traffic flow patterns.  Such prediction is possible by analyzing the real-time transportation data from correlative roads and vehicles.  This article first gives a brief introduction to the transportation data, and surveys the state-of-the-art prediction methods.  Then, we verify whether or not the prediction performance is able to be improved by fitting actual data to optimize the parameters of the prediction model which is used to predict the traffic flow.  Such verification is conducted by comparing the optimized time series prediction model with the normal time series prediction model.  This means that in the era of big data, accurate use of the data becomes the focus of studying the traffic flow prediction to solve the congestion problem.  Finally, experimental results of a case study are provided to verify the existence of such performance improvement, while the research challenges of this data-analytics-based prediction are presented and discussed.
\end{abstract}

\IEEEpeerreviewmaketitle


\section{Introduction}
\label{sec:introduction}
\begin{figure*}[!ht]
  \centering
  \includegraphics[width=7in]{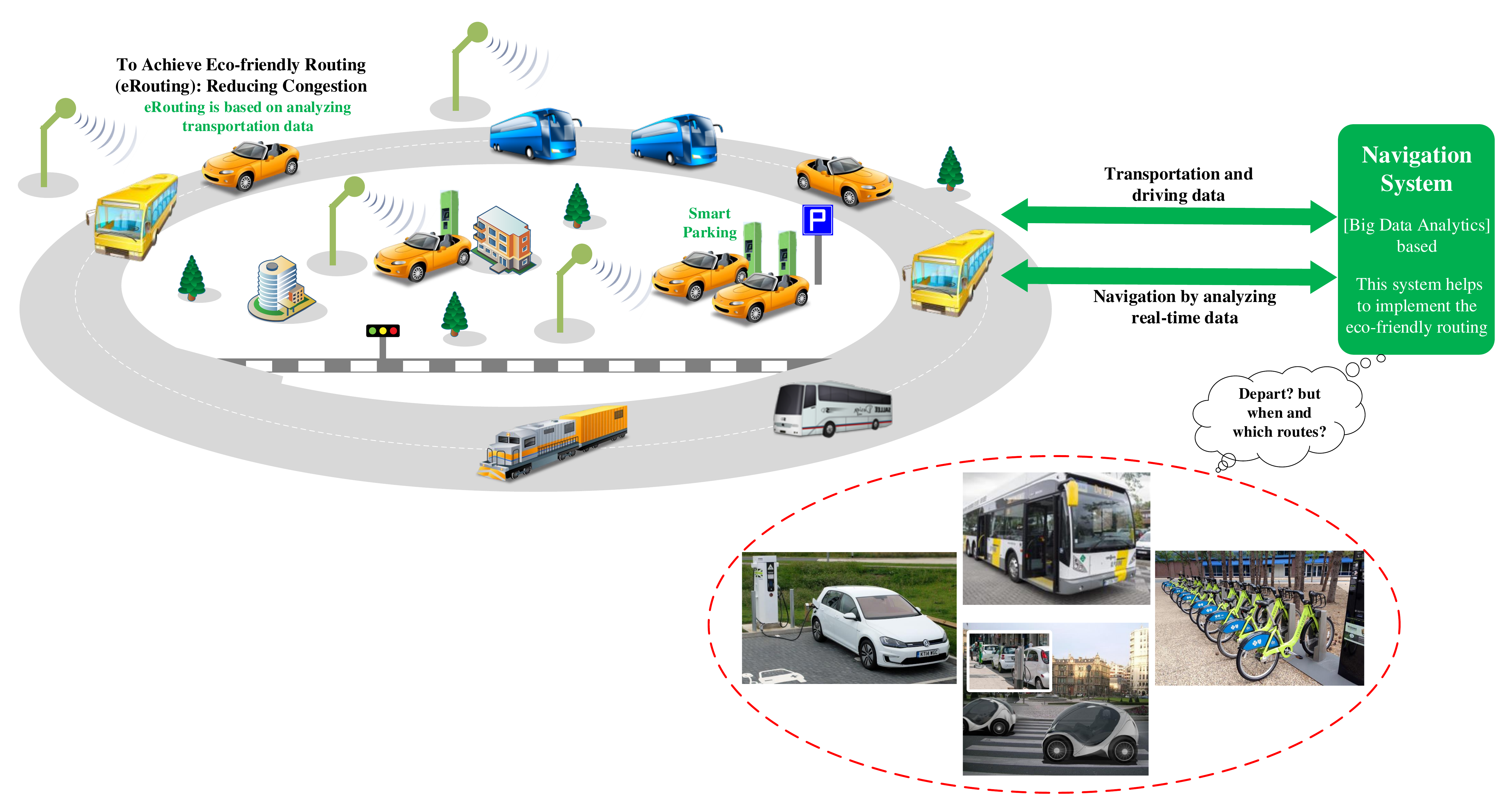}\\
  \caption{Traffic congestion avoidance.  This scenario shows how to avoid congestion by analyzing transportation data, where the data comes from? where does the data go? how the data is used? and what can be achieved by analyzing the data?}
  \label{fig:scenario}
\end{figure*}

Traffic flow prediction is an important research issue in an Intelligent Transportation System (ITS), and it can be used as an important metric to solve the traffic congestion problem.  Traffic congestion is considered a serious problem in big cities around the world.  In a study of 471 U.S. urban areas in 2014~\cite{scorecard2015}, the extra energy cost due to the traffic congestion was estimated at \$160 billion (3.1 billion gallons of fuel).  In addition, long periods of traffic congestion force the release of more carbon dioxide (CO$_{2}$) greenhouse gases into the atmosphere and increase the number of accidents.  This can in turn produce severe public health risks which will dramatically increase medical treatments' costs.  Predicting traffic flow patterns can help reduce traffic congestion and therefore reduce the amount of CO$_{2}$ emissions as well as save lives.  A scenario of using traffic flow prediction to avoid traffic congestion is illustrated in Figure~\ref{fig:scenario}.

Figure~\ref{fig:scenario} illustrates a scenario: how to avoid congestion in an Intelligent Transportation System (ITS), and what can be achieved by avoiding congestion?  In this scenario, the transportation and driving data are collected from various devices of the ITS.  This data is then analyzed to help the navigation of vehicles.  Meanwhile, other vehicles that are preparing to hit the road want to know the current traffic conditions to make a decision of: (i) whether to start the journey and when? and (ii) which routes can be used and which is the best?  On this basis, the congestion can be reduced to achieve eco-friendly routing (low-carbon transportation).

\begin{figure}[!ht]
  \centering
  \includegraphics[width=2.5in]{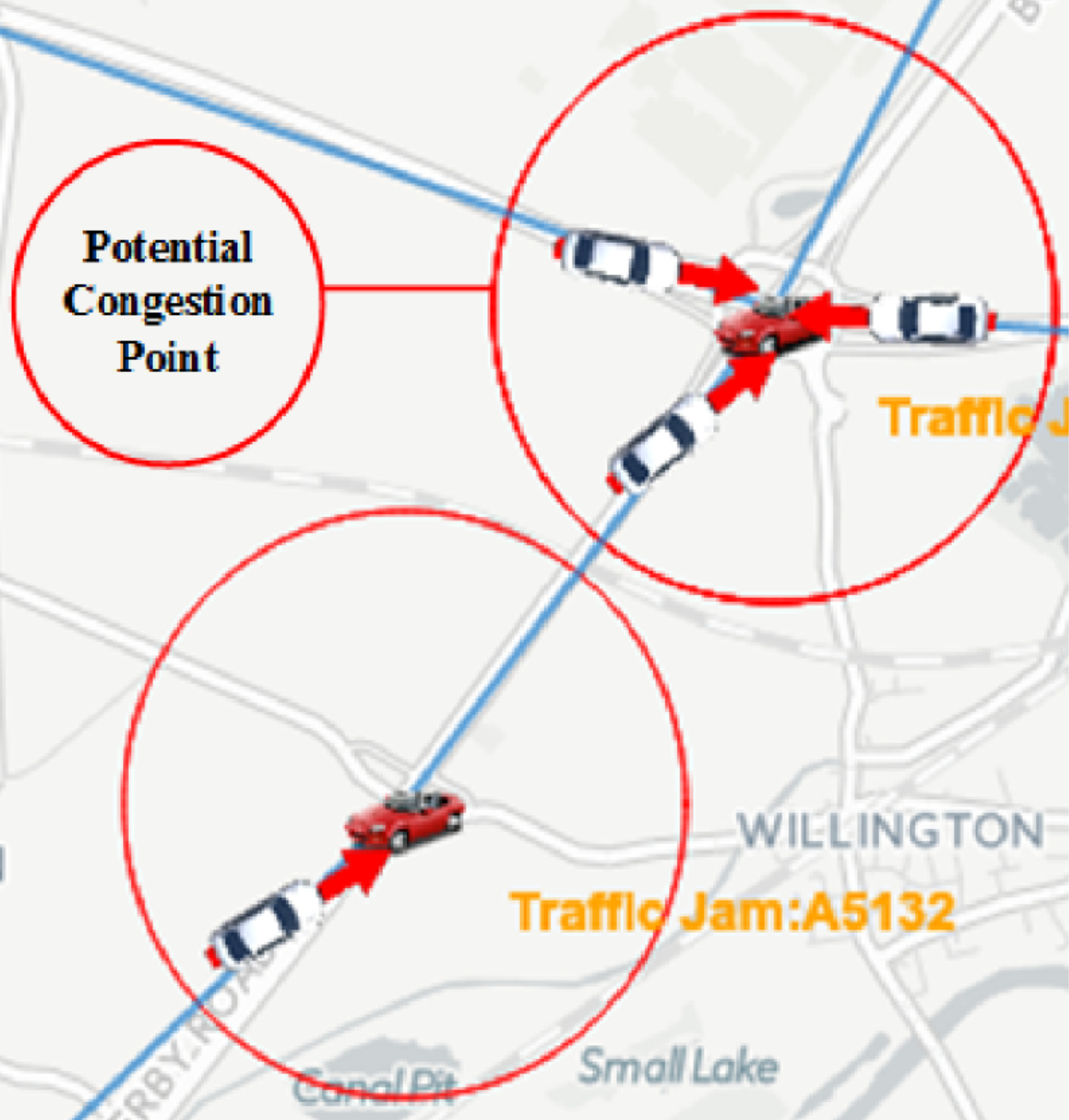}\\
  \caption{In this scenario, the traffic flow in an intersection is predictable.  The traffic flow of the intersection comes from the traffic roads connecting at the intersection.}
  \label{fig:problem_description}
\end{figure}

The traffic flow is predictable by analyzing the relevance of traffic conditions between different traffic roads, as shown in Figure~\ref{fig:problem_description}.

Figure~\ref{fig:problem_description} illustrates the possibility of knowing whether or not there is a potential congestion point in advance.  Such knowledge is acquired by predicting the traffic flow for the specific route.  This is due to the continuity of traffic flow among different traffic roads.  It means that the flow of a traffic road comes from other roads that connect to it.

It is possible to analyze the relevance of the traffic conditions between different traffic roads by tracing the change of the traffic flow, and such tracing can be achieved by analyzing the real-time transportation data submitted by the wireless devices embedded into the vehicles and road side units (RSUs) installed in the correlative roads~\cite{al2015internet}.

The trajectory data of vehicles (a kind of transportation data) is the most commonly used by the traffic flow prediction.  Trajectories provide important information on the mobility of vehicles, where the moving pattern of vehicles characterizes the traffic flow of a transportation system.  The trajectory data of vehicles is becoming easily available in current transportation systems due to the prevalence of the global positioning system (GPS) and other localization technologies.  A trajectory generated by a moving vehicle is usually described by a temporal sequence of spatial locations with their timestamps.  These trajectories convey underlying information about how the traffic flow of the transportation system changes.

\begin{figure}[!ht]
  \centering
  \includegraphics[width=3in]{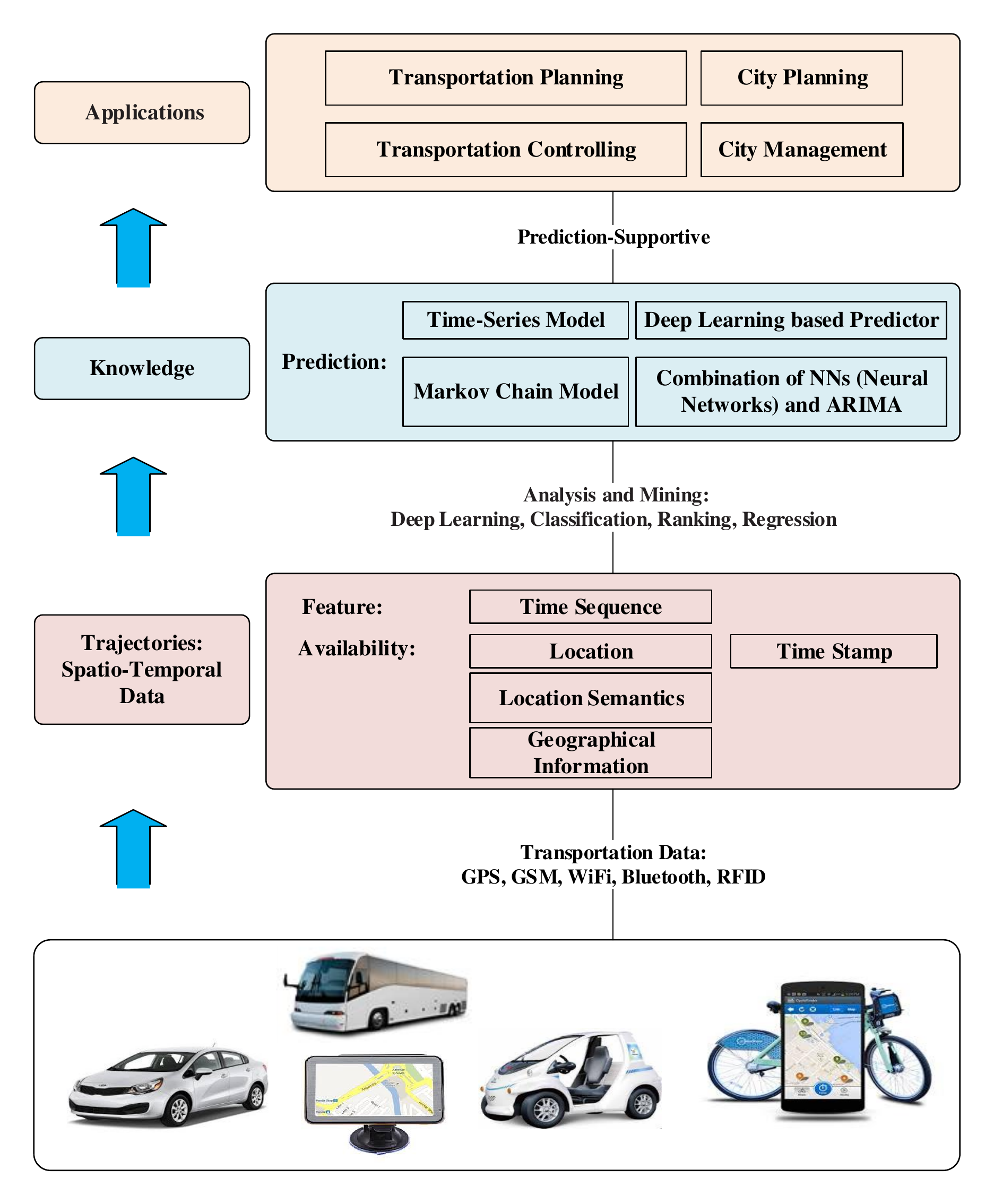}\\
  \caption{General framework of data analytics based traffic flow prediction.}
  \label{fig:general_framework}
\end{figure}

From the traffic congestion problem to traffic flow prediction, and then further to this kind of transportation data: trajectory, the research issue becomes how to analyze and mine the trajectory data to solve the congestion problem.  Analyzing and mining trajectory data can extract and reveal inherent information or knowledge about potential congestion.  It will benefit broad applications, for example, (i) it is possible to reduce exhaust emission by congestion knowledge based path planning to enhance the level of public health; (ii) improving the public security in transportation systems is possible by avoiding the traffic accidents caused by congestion.  Figure~\ref{fig:general_framework} illustrates the general framework of such data analytics based traffic flow prediction. 

The general framework consists of three modules: (i) data collection module: transportation data is able to be collected from vehicle-mounted GPS, GSM, WiFi, Bluetooth and RFID, for example, sharing bikes can be embedded GPS, GSM, WiFi, Bluetooth, and even RFID to track and record: when and where the bikes are shared, and the traveling trajectories of these bikes.  The trajectory is a kind of spatio-temporal data, and this kind of data is with the time sequence feature.  In such data, this information is available: location, location semantics, geographical information and time stamp; (ii) data analytics based prediction module: the collected transportation data is analyzed by deep learning, classification, ranking and regression to acquire useful knowledge as the basis of traffic flow prediction.  These four technologies are able to be used to conduct the data analytics based prediction: time-series model, deep learning based predictor, markov chain model, and combination of Neural Networks (NNs) and Autoregressive Integrated Moving Average (ARIMA); and (iii) application module: the predicted results are able to be used to support many special applications to improve the life experience in cities, for example, transportation planning, transportation controlling, city planning, and city management.


\section{Transportation Data}
\label{sec:traffic_data}
What is transportation data and what kind of transportation data is used in this article?

\textbf{What is transportation data?}  Transportation data is a kind of data that describes the information related to transportation systems, and it contains various vehicle and road information, e.g., the trajectory and speed information of vehicles, and the length information of traffic roads.  It is helpful to improve the performance of transportation related applications by analyzing such data, for example, (i) traffic flow prediction; it is the attempt to estimate the number of vehicles that will travel on the traffic roads of a transportation system in the future.  This prediction enables us to understand and develop an optimal transportation system with efficient movement of traffic.  It will also allow us to minimize the traffic congestion by designing optimal overall route planning for vehicles based on the predicted traffic flow.  The flow of a traffic road has a connection with the flows of prior roads which are connected to this traffic road.  It means that it is possible to predict the flow of the current traffic road by calculating the potential fractional flows from the prior roads and the previous time period.  The potential fractional flow is able to be calculated by analyzing the transportation data (the trajectory and speed data of vehicles); (ii) transportation planning; it is the process of defining the future design for transportation systems to prepare for the future needs on moving people and goods to destinations.  In this application, it is possible to acquire the underlying knowledge of transportation systems by mining the transportation data from the systems.  With the acquired knowledge, the transportation planning will be more reasonable.

\begin{figure}[!ht]
  \centering
  \includegraphics[width=3.5in]{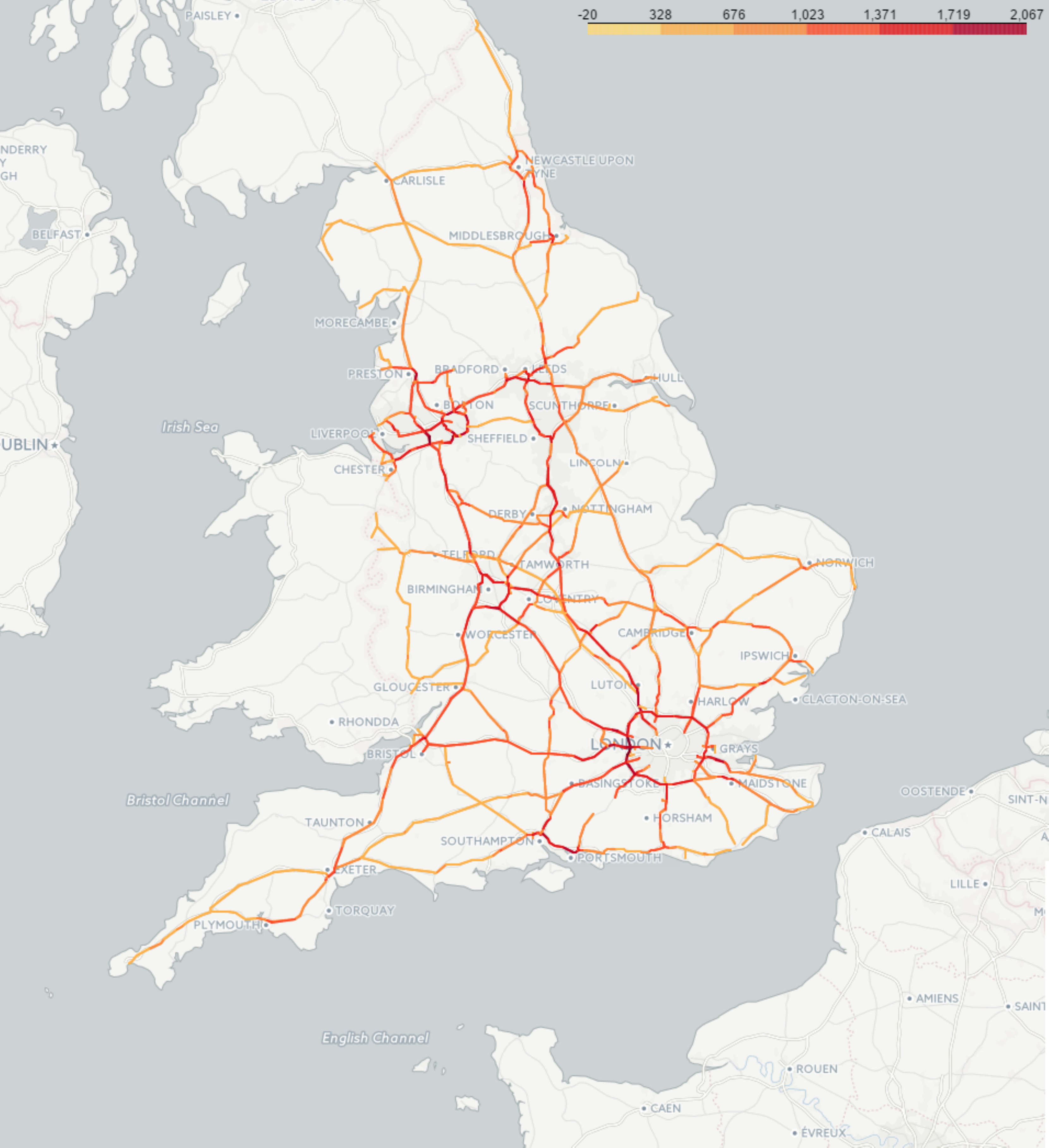}\\
  \caption{Flow information of traffic roads is acquired by analyzing the trajectory data of the vehicles driving on these traffic roads.  The update period of this information is 15 minutes.  Highways and arterial roads are marked by different color depths, and the deeper color denotes the heavier traffic load on the corresponding road.}
  \label{fig:data_description}
\end{figure}

\textbf{What kind of transportation data is used in this article?}  In this study, trajectory data of vehicles is used, which is a kind of series transportation data.  This data can be used to calculate and predict the traffic flow of a transportation system.  It is collected from the transportation system in England.  This transportation system has 2501 traffic roads\footnote{A two-way highway/arterial road is counted as two traffic roads.} covering 300 miles of England highways and arterial roads, which is illustrated in Figure~\ref{fig:data_description}.

The important part of trajectory data is the location information.  With the location information, trajectory data can provide actual semantics to make such data be used by traffic flow prediction.  Flow information of a traffic road is acquired by counting the number of vehicles being driven on the current road.  So in trajectory data, the location information is required to filter out the vehicle driving records on the current road rather than other roads.  Moreover, for traffic flow prediction, the traffic flow of the current road converges from the traffic flow of prior roads.  So the location information becomes important to track the traffic flow changes.  Trajectory data relies on localization technology to provide location information.  The commonly used measurement methods of localization to generate location-containing trajectory data include GPS, WiFi, GSM, Bluetooth, and RFID.


\section{State-of-the-art Prediction Methods}
\label{sec:state-of-the-art_prediction_methods}
Many kinds of techniques have been proposed to address the traffic flow prediction problem.  The details are summarized in the following literature list. 
\begin{table*}[!ht]
\centering
\label{tab:techniques_traffic_flow_prediction}
\begin{tabular}{ll}
\bottomrule
  Techniques & Typical Works and Advantages\\\midrule
  \cellcolor{black!15}Bayesian Networks
  & \begin{tabular}[t]{
     >{\ttfamily\raggedright}p{10cm}
     >{\sffamily\raggedright}p{10cm}
     >{\sffamily}p{\dimexpr\textwidth-12\tabcolsep-5\fboxsep-7.5cm\relax}
  }
  \cite{sun2005traffic} this work presented a spatio-temporal Bayesian network predictor.  This work employed the Gaussian Mixture Model and the Competitive Expectation Maximization algorithm in order to approximate the joint probability distribution of the nodes in the Bayesian network.
  \end{tabular} \\[.5ex]
  \cline{2-2}

  \cellcolor{black!15} & \begin{tabular}[t]{
     >{\ttfamily\raggedright}p{10cm}
     >{\sffamily\raggedright}p{10cm}
     >{\sffamily}p{\dimexpr\textwidth-12\tabcolsep-5\fboxsep-7.5cm\relax}
  }
  \cite{sun2006bayesian} this work proposed a model: (i) this model has the ability to create a Bayesian network to the traffic flow on a given road at a given time; (ii) this model has an ability to cope with the data-incomplete problem.
  \end{tabular} \\[.5ex]\cline{2-2}

  \cellcolor{black!15} & \begin{tabular}[t]{
     >{\ttfamily\raggedright}p{10cm}
     >{\sffamily\raggedright}p{10cm}
     >{\sffamily}p{\dimexpr\textwidth-12\tabcolsep-5\fboxsep-7.5cm\relax}
  }
  \cite{castillo2008predicting} the technique of the traffic prediction, which was presented in this work, used the capability of traffic flows to reproduce the real behavior of vehicles and to construct the Gaussian Bayesian network employing the special characteristics of the variables of the traffic flow.
  \end{tabular} \\[.5ex]\cline{2-2}

  \cellcolor{black!15} & \begin{tabular}[t]{
     >{\ttfamily\raggedright}p{10cm}
     >{\sffamily\raggedright}p{10cm}
     >{\sffamily}p{\dimexpr\textwidth-12\tabcolsep-5\fboxsep-7.5cm\relax}
  }
  \cite{zhu2016short} this work utilized a linear conditional Gaussian Bayesian Network (BN) model to consider the impacts of both spatial and temporal dimensions of traffic as well as speed information for short-term traffic flow prediction.
  \end{tabular} \\[.5ex]\midrule

  Neural Networks
  & \begin{tabular}[t]{
     >{\ttfamily\raggedright}p{10cm}
     >{\sffamily\raggedright}p{10cm}
     >{\sffamily}p{\dimexpr\textwidth-12\tabcolsep-5\fboxsep-7.5cm\relax}
  }
  \cite{chan2012neural} this work proposed a method to make short-term prediction based on Neural Networks (NNs).
  \end{tabular} \\[.5ex]\cline{2-2}

  & \begin{tabular}[t]{
     >{\ttfamily\raggedright}p{10cm}
     >{\sffamily\raggedright}p{10cm}
     >{\sffamily}p{\dimexpr\textwidth-12\tabcolsep-5\fboxsep-7.5cm\relax}
  }
  \cite{chan2013road} this work provided a traffic flow predictor based on the architecture of fuzzy neural networks.
  \end{tabular} \\[.5ex]\midrule

  \cellcolor{black!15}Time-Series Models
  & \begin{tabular}[t]{
     >{\ttfamily\raggedright}p{10cm}
     >{\sffamily\raggedright}p{10cm}
     >{\sffamily}p{\dimexpr\textwidth-12\tabcolsep-5\fboxsep-7.5cm\relax}
  }
  \cite{calheiros2015workload} the statistical time-series model, ARIMA is fitted to time series data either to better understand the data or to predict future points in the series.  The ARIMA model is not suited for the prediction which is based on this kind of data: (i) there is the information missing problem in the data, and (ii) the filling of data is problematic as the situation is complex.
  \end{tabular} \\[.5ex]\cline{2-2}

  \cellcolor{black!15} & \begin{tabular}[t]{
     >{\ttfamily\raggedright}p{10cm}
     >{\sffamily\raggedright}p{10cm}
     >{\sffamily}p{\dimexpr\textwidth-12\tabcolsep-5\fboxsep-7.5cm\relax}
  }
  \cite{williams2003modeling} this work presented the theoretical basis for modelling univariate traffic conditions as Seasonal Autoregressive Integrated Moving Average (SARIMA) processes.  Fitted SARIMA models provide equations which can be used to produce single and multiple interval prediction.
  \end{tabular} \\[.5ex]\cline{2-2}

  \cellcolor{black!15} & \begin{tabular}[t]{
     >{\ttfamily\raggedright}p{10cm}
     >{\sffamily\raggedright}p{10cm}
     >{\sffamily}p{\dimexpr\textwidth-12\tabcolsep-5\fboxsep-7.5cm\relax}
  }
  \cite{stathopoulos2003multivariate} the focus in this work was on producing time-series state space models which are flexible and explicitly multivariate. The models enable to jointly consider the data from different detectors, and are able to model a wide variety of univariate models, for instance, ARIMA.  The results were compared to those achieved from the ARIMA model and were found to be superior.
  \end{tabular} \\[.5ex]\midrule

  Kalman Filtering Theory
  & \begin{tabular}[t]{
     >{\ttfamily\raggedright}p{10cm}
     >{\sffamily\raggedright}p{10cm}
     >{\sffamily}p{\dimexpr\textwidth-12\tabcolsep-5\fboxsep-7.5cm\relax}
  }
  \cite{guo2014adaptive} this work investigated the performance of the adaptive Kalman Filter for short-term traffic flow prediction.  Moreover, this work presented the empirical results from the application of the adaptive Kalman Filter to real-world data.  Kalman Filtering is applicable to short-term stationary or non-stationary stochastic phenomena and it yields good traffic prediction accuracy.
  \end{tabular} \\[.5ex]\midrule

  \cellcolor{black!15}Markov Chain Model
  & \begin{tabular}[t]{
     >{\ttfamily\raggedright}p{10cm}
     >{\sffamily\raggedright}p{10cm}
     >{\sffamily}p{\dimexpr\textwidth-12\tabcolsep-5\fboxsep-7.5cm\relax}
  }
  \cite{yu2003short} this work modelled the traffic flow as a high-ordered Markov Chain.  The method proposed in this work employs current and recent values of the traffic flow, and describes the future value.  This future value is assumed that the predicted state has a probability distribution, and both current state and most recent states determine the next state.
  \end{tabular} \\[.5ex]\midrule

  Combination of NNs and ARIMA
  & \begin{tabular}[t]{
     >{\ttfamily\raggedright}p{10cm}
     >{\sffamily\raggedright}p{10cm}
     >{\sffamily}p{\dimexpr\textwidth-12\tabcolsep-5\fboxsep-7.5cm\relax}
  }
  \cite{fard2014hybrid} this work proposed a hybrid model combining NNs and ARIMA, which is capable of exploiting the strengths of traditional time series approaches and artificial neural networks.
  \end{tabular} \\[.5ex]\midrule

  \cellcolor{black!15}Deep Learning based Predictor
  & \begin{tabular}[t]{
     >{\ttfamily\raggedright}p{10cm}
     >{\sffamily\raggedright}p{10cm}
     >{\sffamily}p{\dimexpr\textwidth-12\tabcolsep-5\fboxsep-7.5cm\relax}
  }
  \cite{lv2015traffic} this work proposed a deep learning based traffic flow prediction method, which considers the spatial and temporal inherent correlations of traffic records.
  \end{tabular} \\[.5ex]
\bottomrule
\end{tabular}
\end{table*}

Many specific methods have been proposed to make prediction in different situations, and it emerges that traditionally there is no single best method for every situation.  It is better to combine several suitable techniques to improve the accuracy of prediction under considering different situations.  It means that the traditional traffic flow prediction methods are not able to satisfy most real-world application requirements.  During the last five years, some studies have tried to use data analytics to solve the traffic flow prediction problem, and the results demonstrate that such schemes are feasible and are able to improve the accuracy of prediction, as the method proposed in~\cite{lv2015traffic}.  Such kind of data analytics based prediction method is able to satisfy the requirements of different applications by analyzing the data from each corresponding specific application.

\section{Performance Comparison}
\label{sec:performance_comparison}
This section compares the performance of two kinds of time series prediction models: (i) optimized time series prediction model, and (ii) normal time series prediction model.  To verify this fact: the prediction performance of a prediction model can be improved by fitting data to optimize model parameters.  ARIMA model is used in this study to make this verification.  Predicting a stationary time series by ARIMA depends on the parameters $(p,d,q)$ of the ARIMA:
\begin{itemize}
  \item The parameter $p$ is the number of Auto-Regressive (AR) terms, for example, if $p$ is 5, the predictors for $x(t)$ will be $x(t-1),...,x(t-5)$;
  \item The parameter $q$ is the number of Moving Average (MA) terms.  MA terms are lagged prediction errors in the prediction equation, for example, if $q$ is 5, the predictors for $x(t)$ will have such lagged prediction errors $e(t-1),...,e(t-5)$, where $e(i)$ is the difference between the moving average and the actual value at the $i^{th}$ instant.
  \item The parameter $d$ is the number of differences when the time series becomes stable.
\end{itemize}

An important concern here is how to determine the values of `p' and `q'.  Determining the values of `p' and `q' will affect the performance of the prediction model.  This study uses Bayesian Information Criterion (BIC) to determine the optimum values of parameters `p' and `q' to avoid over-fitting, when fitting a model by training data.  BIC is a criterion for model selection among a finite set of models, and the model with the lowest BIC value is preferred.  It is based, in part, on the likelihood function.

\begin{figure*}
    \centering
    \begin{subfigure}[b]{0.45\textwidth}
        \includegraphics[width=\textwidth]{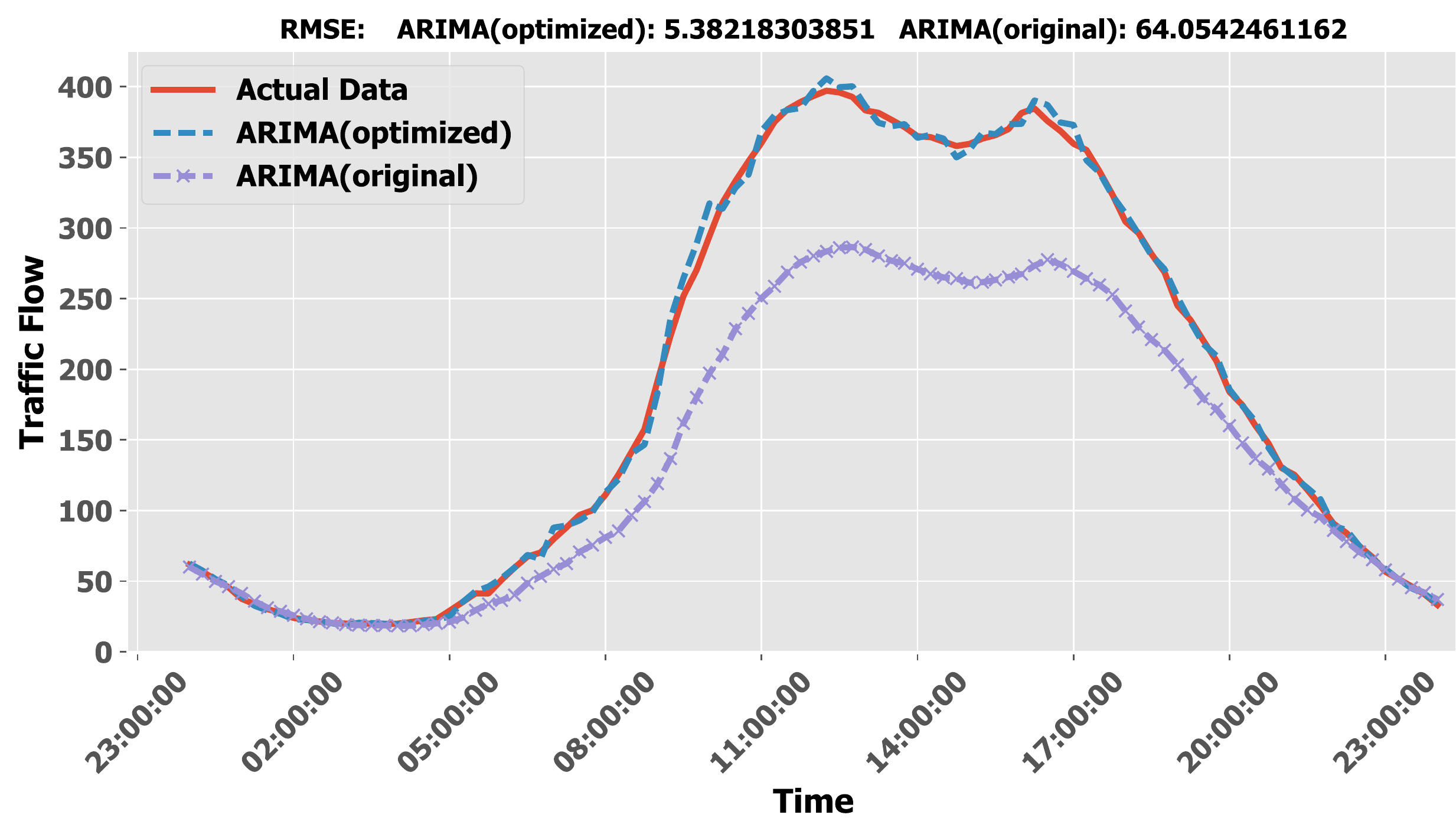}
        \caption{Actual traffic flow data of March 15th, 2015, and the predicted results by normal and optimized ARIMA models.}
        \label{fig:15thMAR}
    \end{subfigure}
    ~~~
    \begin{subfigure}[b]{0.45\textwidth}
        \includegraphics[width=\textwidth]{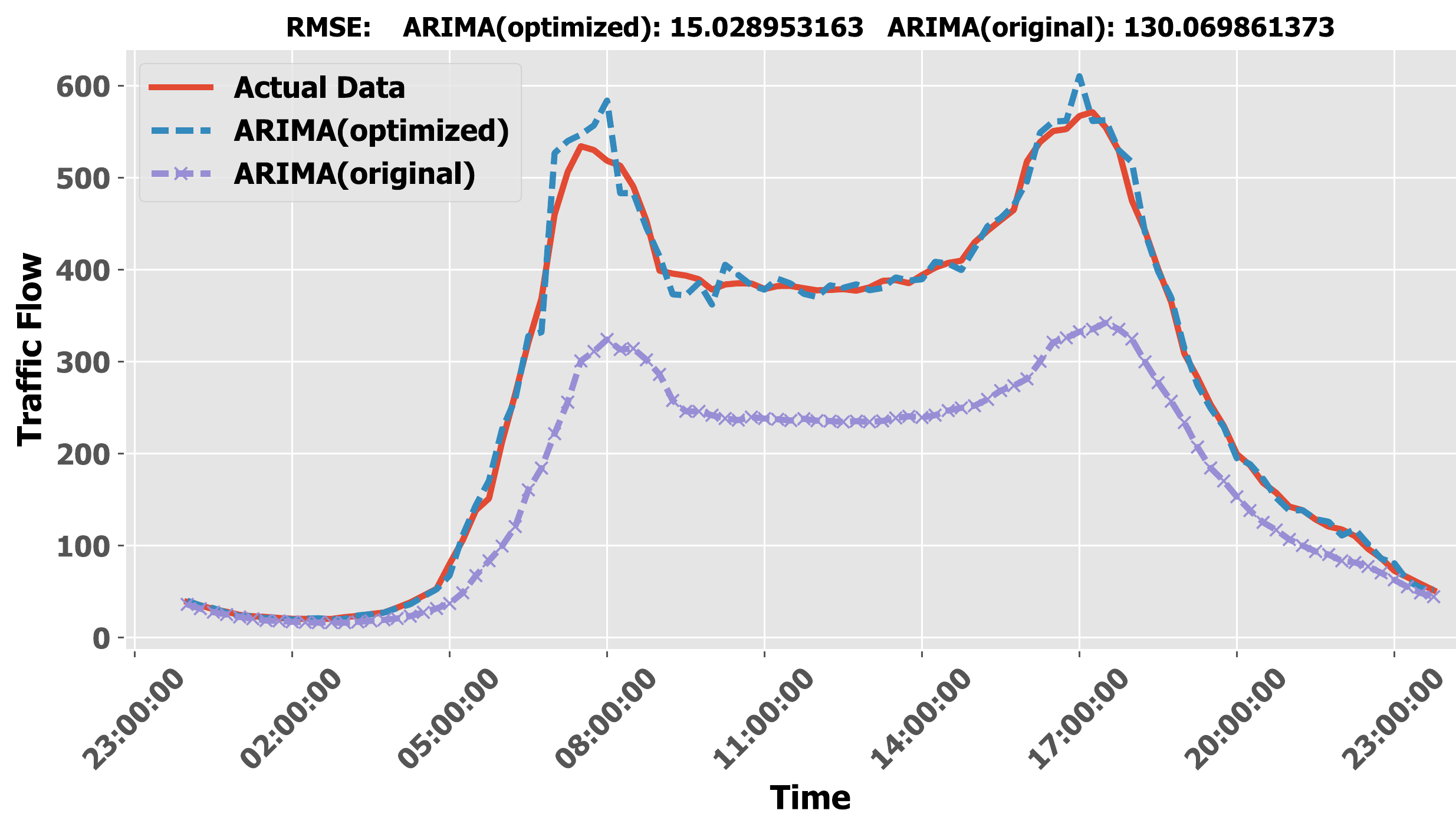}
        \caption{Actual traffic flow data of March 31st, 2015, and the predicted results by normal and optimized ARIMA models.}
        \label{fig:31stMAR}
    \end{subfigure}
    \begin{subfigure}[b]{0.45\textwidth}
        \includegraphics[width=\textwidth]{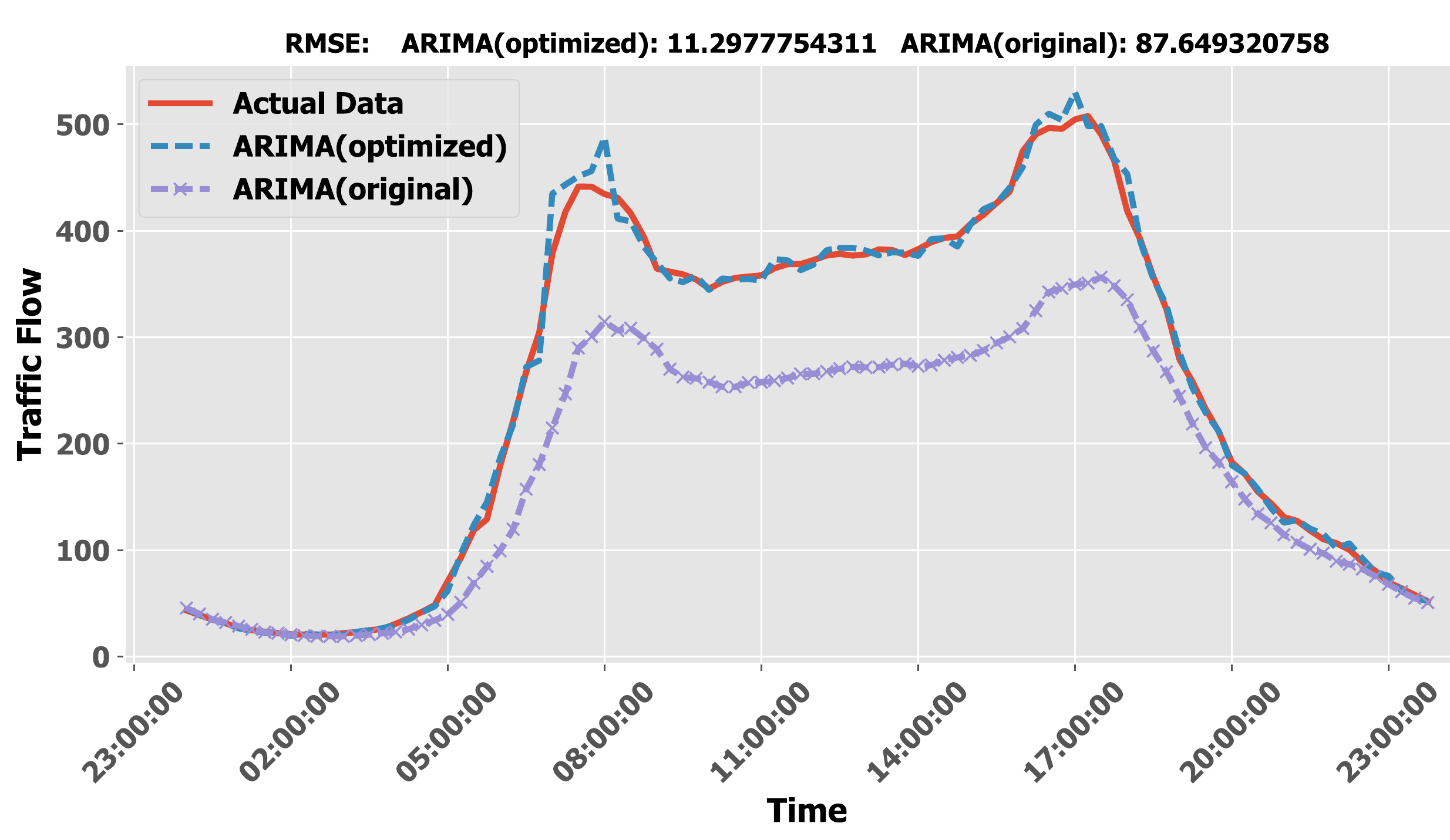}
        \caption{Average results for 31 days of March: actual traffic flow and the predicted results by normal and optimized ARIMA models.}
        \label{fig:avg}
    \end{subfigure}
    \caption{Comparing with the normal ARIMA, optimized ARIMA achieves better prediction results in the aspect of the changing trend and values of traffic flow.  Root-Mean-Square Errors (RMSE) are marked in each subfigure to indicate the prediction performance of the normal and optimized ARIMA models.}
    \label{fig:comparison_results}
\end{figure*}

Comparative results are illustrated in Figure~\ref{fig:comparison_results}.  Each line is the average of 2501 traffic roads.

The RMSE of a model with respect to the ``Traffic Flow'' is defined as: the square root of the mean square error between the values actually observed and the values predicted by a model.  The RMSE estimated from the optimized ARIMA is less than the estimated result from the normal ARIMA.  The prediction accuracy of the time series model ARIMA is able to be obviously improved by optimizing relevant parameters.

Figure~\ref{fig:15thMAR} and Figure~\ref{fig:31stMAR} illustrate the comparative results with the data of two selected dates, March 15th and 31st, 2015.  For the RMSE, the optimized ARIMA is much better than the normal ARIMA.  Figure~\ref{fig:avg} provides the average results for 31 days of March, 2015.  On these basis, it is observed that improving the performance of a prediction model is possible by optimizing model parameters with analyzing sequential transportation data.


\section{Research Challenges}
\label{sec:research_issues}
\begin{figure}[!ht]
  \centering
  \includegraphics[width=3.6in]{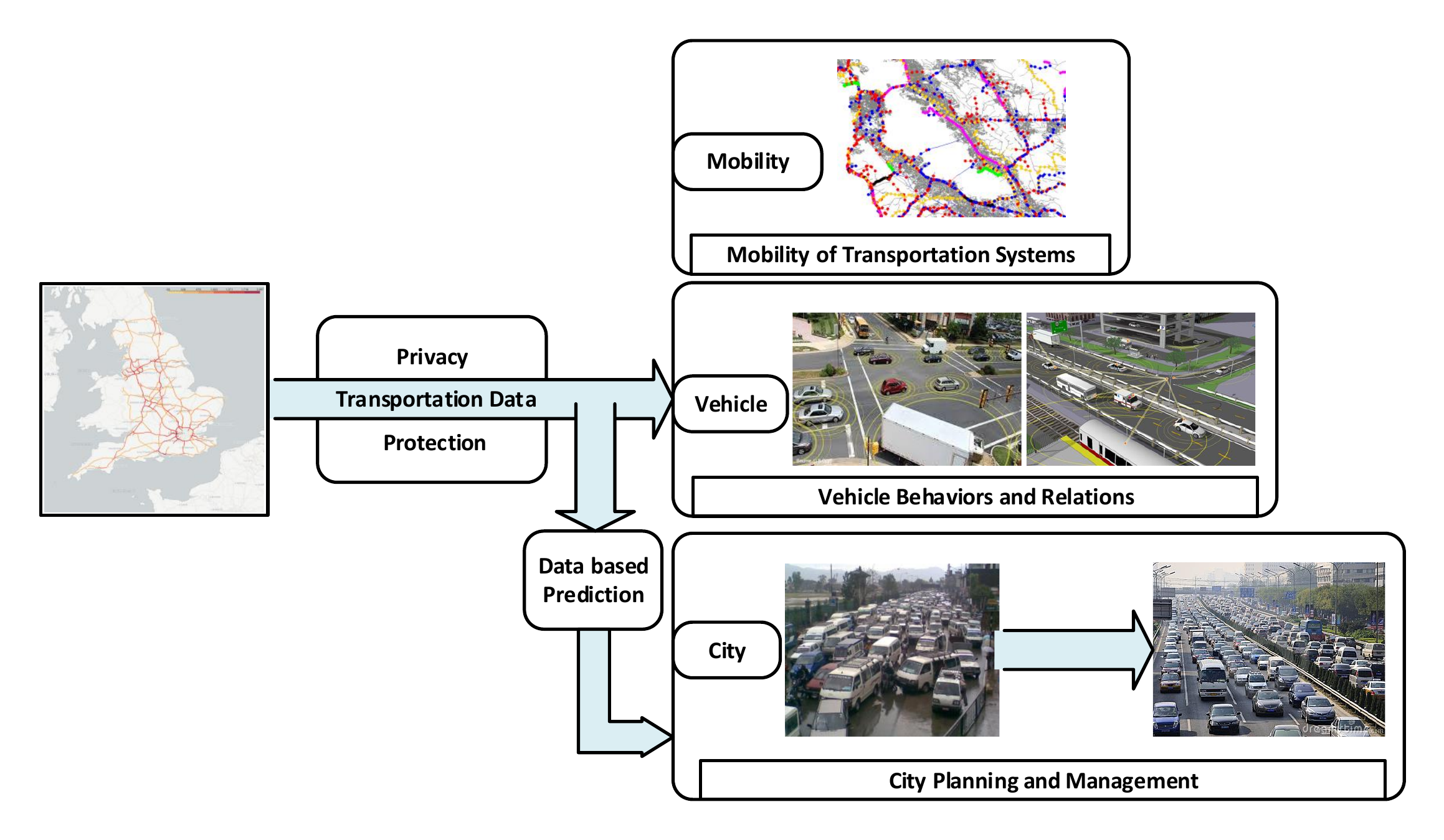}\\
  \caption{Research challenges of data supportive prediction for traffic flow.}
  \label{fig:research_issues}
\end{figure}

The pattern prediction of traffic flow, which is supported by data analytics, strongly relies on the transportation data which records the vehicle mobility of transportation systems.  Moreover, transportation data implies the underlying patterns and laws of the vehicle mobility.  It means that the data is embedded with the rich information of the traffic dynamics of a city.  Meanwhile, transportation information may cause privacy issues.  This article classifies the research challenges of data analytics supportive traffic flow prediction, as shown in Figure~\ref{fig:research_issues}.
There are two important aspects for the research challenges of this kind of prediction: data and data based prediction.

\textbf{Data:}
\begin{enumerate}
  \item [i)] \emph{Privacy Protection:} it is necessary and important to protect the personal privacy information which is contained in transportation data.  Such information includes the locations, driving trajectories, properties and vehicles' plate numbers.  It is a worthy research challenge to protect this information when the transportation data is analyzed to obtain useful information and knowledge.
  \item [ii)] \emph{Mobility:} in transportation systems, understanding and predicting the patterns of mobility is the basis of studying and solving a variety of practical problems, e.g., traffic congestion problems.  The mobility produces large amounts of spatio-temporal data, and this kind of data is accompanied with time and location information.  It is a worthy research challenge to analyze the spatio-temporal data for learning and predicting the patterns of mobility.  Moreover, traffic flow is an important reflection for the mobility in transportation systems.  Therefore, traffic flow prediction is also an important and worthy research challenge.  Analyzing the spatio-temporal data is another research challenge as the basis of understanding, learning and predicting the patterns of mobility and traffic flow.  Spatio-temporal data is a kind of time series data.  To analyze this kind of data, a time series analysis model is necessary, and for different applications, the objective functions are different because of the different requirements, and how to combine a time series analysis model with an objective function for special requirements is an important challenge that requires further investigation.
  \item [iii)] \emph{Vehicle:} the vehicles in a transportation system produce large amounts of trajectory data.  This kind of data can be used to mine vehicle behaviors and relations.  If we know the behaviors of vehicles and the relations between these vehicles, it is possible to predict the change of traffic flow and then to avoid congestion by traffic control and management.  For example, if there is congestion at a certain intersection and the relevant vehicles have the same destination, these vehicles have great possibility to use the same route, and the congestion will happen again at the next intersection, if there are no control measures in advance.
\end{enumerate}

\textbf{Big Data:}
In these applications, ``big data'' can be achieved whether from ``mobility'' and/or from ``vehicles'': (i) the mobility produces large amounts of spatio-temporal data, and (ii) the vehicles produce large amounts of trajectory data.  Such ``large amounts of'' is a kind of ``big'', and for analyzing this kind of data, big data analytics is needed.  Such analytics is different from traditional data analysis.  For example, deep learning and NNs can be used to analyze such big data, but how to learn such data to extract useful information and knowledge, and learning for what, are worthy challenges which need to be studied.

\textbf{Wireless Big Data:}
Most of the big data of an intelligent transportation system (the trajectory data of vehicles) is collected using wireless devices.  These devices can be wireless sensor nodes which are embedded on the traffic road surfaces to detect passing vehicles.  There will have data missing problems in this kind of wireless data because of some unpredictable problems, e.g., sensor node failure.  In some scenarios, if the critical data is missing or has errors, it will make the analysis results gravely deviate from making correct conclusions.  For example, if the data from a device is not correct and the data is used to perform analysis, the result will be incorrect and cannot be used as useful information or knowledge to aid decision making.  Ensuring the integrity of the wireless big data information is still a great challenge that needs to be addressed by experts in the field.

\textbf{City Data Based Prediction:}
The prediction results in analyzing transportation data can help in city planning and management.  The large trajectory data produced by moving vehicles can help predict the traffic flow that in turn can greatly improve lives.  In addition, this can motivate the design of intelligent traffic lights that can achieve automatic traffic flow global scheduling.  Such scheduling is based on the understanding and prediction of global traffic conditions.  So, it is important to know whether the current traffic flow will impact roads during the next time period or not.  These issues are still a challenge that need researchers' attention in the near future.


\section{Conclusion}
\label{sec:conclusions}
Transportation data is the sampling of dynamics of moving objects in the temporal and spatial dimensions.  Analysis and mining of such time series data is becoming a promising way to discover the underlying knowledge on vehicular activities, vehicles' relations and even city dynamics.  It helps to well understand and predict the pattern of traffic flow which is an important aspect of traffic dynamics.  Traffic flow prediction can be exploited in a wide range of potential applications to make a city smarter, safer, more livable, and can help reduce congestion and pollution.  However, it remains challenging to analyze time series data to acquire useful knowledge for problem solving, because of data heterogeneity and data incompleteness in a complex and dynamic transportation system.  So, in this article we have introduced the transportation data.  Then, we discussed state-of-the-art prediction methods.  We run some experiments in order to compare the optimized time series prediction model with the normal time series prediction model.  Our results show that the prediction performance is able to be improved by fitting actual data to optimize the parameters of the prediction model.  In addition, we have discussed typical techniques to address the traffic flow prediction problem.  Finally, we listed some challenges that will provide leads to future researchers in this area.




\bibliographystyle{IEEEtran}
\bibliography{IEEEfull}


\end{document}